\pdfoutput=1
\documentclass[11pt]{article}

\usepackage[]{acl}

\usepackage{times}
\usepackage{latexsym}
\usepackage{amsmath}
\usepackage{color}
\newcommand{\todo}[1]{}
\renewcommand{\todo}[1]{{\color{red} TODO: {#1}}}

\usepackage{graphicx}
\usepackage{multirow}
\usepackage[ruled,vlined]{algorithm2e}
\DeclareMathOperator*{\argtop}{argtop}

\usepackage[T1]{fontenc}
\usepackage[utf8]{inputenc}

\usepackage{microtype}

\title{On the Complementarity of Data Selection\\ and Fine Tuning for Domain Adaptation}

\author{Dan Iter \\
  Stanford University \\
  \texttt{daniter@stanford.edu} \\\And
  David Grangier \\
  Google Brain \\
  \texttt{grangier@google.com} \\}

\begin{document}
\maketitle
\begin{abstract}
Domain adaptation of neural networks commonly relies on three training phases: pretraining, selected data training and then fine tuning. 
Data selection improves target domain generalization by training further on pretraining data identified by relying on a small sample of target domain data. This work examines the benefit of data selection for language modeling and machine translation. Our experiments assess the  complementarity of selection with fine tuning and result in practical recommendations: 
(i) selected data must be similar to the fine-tuning domain but not so much as to erode the complementary effect of fine-tuning;
(ii) there is a trade-off between selecting little data for fast but limited progress or much data for slow but long lasting progress;
(iii) data selection can be applied early during pretraining, with performance gains comparable to long pretraining session; 
(iv) data selection from domain classifiers is often more effective than the popular contrastive data selection method.
\end{abstract}
\graphicspath{ {img/} }

\section{Introduction}
\label{sec:introduction}

Machine learning models, and neural networks in particular, benefit from large training sets. However, for many application domains, the amount of training data representative of the inference conditions is limited. It is therefore common to train a model over a large amount of generic, out-of-domain data while relying on a small amount of target 
domain data to adapt such a model. In the recent years, a large body of work has
focused on leveraging large amount of web data to train neural networks for language modeling~\cite{peters-etal-2018-deep,devlin-etal-2019-bert} or translation 
systems~\cite{banon-etal-2020-paracrawl,koehn-etal-2020-findings}. 
Such systems are then adapted to the target distribution, typically via fine tuning \cite{liu2019roberta,raffel-t5-2020}. 
This work studies data selection, an intermediate training phase that visits a subset of the out-of-domain data that is deemed closer to the target domain.

% present domain classifiers for NMT
% introduce complementarity
Previous work has proposed conducting a data selection step after pretraining~\cite{vanderwees2017dynamic,wang-etal-2018-denoising,gururangan-etal-2020-dont,aharoni-goldberg-2020-unsupervised}, either as a final training stage or before regular fine tuning. 
%Like fine tuning, data selection relies on a small set of data from the target domain. 
%It uses this data to identify a subset of the out-of-domain pretraining set which might be the most helpful to improve generalization on the target distribution. 
Data selection is meant to identify a subset of the out-of-domain pretraining set which might be the most helpful to improve generalization on the target distribution.
This selection is typically conducted by estimating the probability that each data point belongs to the target domain~\cite{moore-lewis-2010-intelligent,axelrod-etal-2011-domain}.
Recently, \cite{aharoni-goldberg-2020-unsupervised} introduced the use of domain classifiers for data selection.

This work examines the benefit of data selection for language modeling and machine translation. We compare different selection methods and examine their effect for short and long pretraining sessions. We also examine the benefit of selecting varying amount of training data and the impact of selection on the subsequent benefit of fine-tuning. In addition to this novel analysis, our machine translation experiments compare the benefit of selecting data with a classifier based on source language, target language or both.

The effectiveness of data selection is dependent on (i) the similarity of the pretraining data to the target domain data, (ii) the precision of the selection method to identify in-domain examples from the pretraining set, (iii) the extent to which training on the selected data is complementary to finetuning. This work focuses on selecting data from the pretraining set so (i) is fixed.
We show that (ii) benefits from the use of domain classifiers, in particular, fine-tuned pretrained language models, outperforming the more popular constrastive methods (eg. \citet{wang-etal-2018-denoising}) in all settings that we tested.
We present the first analysis of (iii), which we refer to as the complimentarity of selected data to finetuning data. We show
that some data selection methods can actually erode the effectiveness of subsequent fine-tuning. In some settings, we even report that a poor complementarity of selection and fine tuning can result in their combination reaching worse results than fine tuning alone. 

Effective application of data selection requires careful selection of when to switch from pretraining to selection, how much selected data to train on and how long to train on selected data before switching to finetuning.
Much of the previous work on data selection either evaluates small models that converge quickly \cite{moore-lewis-2010-intelligent,axelrod-etal-2011-domain} or does not describe the extent of grid search over selection size, number of steps of pretraining and number of steps of training on selected data.
We are the first to analyze the hyperparameter selection tradeoffs for data selection on large neural models, where models may be undertrained \cite{liu2019roberta} and evaluating many selection sizes may be prohibitively expensive.
We evaluate data selection on checkpoints with variable numbers of pretraining steps and show that data selection provides consistent results between minimally and extensively pretrained models.
We also show the challenges of searching over selection sizes because smaller selection sizes always converge more quickly but are outperformed by larger selection sizes trained for more steps.

Our findings are the following: 
(i) the data selection mechanism must select data that is similar, but complementary to the fine tuning dataset
(ii) the amount of selected data introduces a trade-off between quick but limited improvements when limiting selection to the best data, and long lasting but slow progress when selecting more data with an overall worse quality,
%(iii) evaluation of data selection must include the computational trade-off between data selection and pretraining for larger models.
%In all settings that we evaluate, incorporating data selection outperforms standard pretraining and fine tuning.
(iii) data selection techniques are not created equal and domain classifiers often outperform contrastive scoring, the most common data selection method,
(iv) we propose three simple variants of domain classifiers for machine translation that can conditions the classifier on either source, target or both. 
%(i) we find that selection methods are beneficial at all stages of pretraining; (ii) we find that selection methods can overfit to the small in-domain training set and that their selection favors examples bringing updates similar to fine tuning, hence impairing the complementary effect of fine tuning after data selection; (iii) we observe that the amount of selected data introduces a trade-off between quick but limited improvements when limiting selection to the best data, and long lasting but slow progress when selecting more data with an overall worse quality. 
We demonstrate these findings on language modeling and two language pairs for neural machine translation.

\section{Related Work}
\label{sec:related_work}

% neural + FT
In Natural Language Processing (NLP), adaptation methods have been applied to language modeling~\cite{moore-lewis-2010-intelligent},
machine translation~\cite{axelrod-etal-2011-domain,daume-iii-jagarlamudi-2011-domain}, dependency parsing~\cite{finkel-manning-2009-hierarchical} or sentiment analysis~\cite{tan2009adapting,glorot2011domain}.
With the growing popularity of neural methods~\cite{collobert11,bahdanau2014neural,goldberg2017neural}, 
the adaptation of neural models via fine tuning has become wide-spread for various NLP applications~\cite{devlin-etal-2019-bert,liu2019roberta, raffel-t5-2020}. Data selection is another popular technique~\cite{van-der-wees-etal-2017-dynamic,wang-etal-2018-denoising}
which can be used on its own or in combination to fine tuning.

% FT
% In fine tuning approaches, a model is first trained on large amount of out-of-domain data during {\it pretraining}. The pretrained model is then fine-tuned over the small target domain dataset, i.e. the loss on the in-domain training data is minimized starting from the weights minimizing the out-of-domain loss from pretraining. This approach is attractive since it allows adapting the same pretrained model to multiple domains. Moreover, fine-tuning is typically short compared to pretraining, which allows exploring adaptation parameters efficiently. Fine-tuning has shown its effectiveness over a wide-range of tasks, ranging from question answering and classification to sentiment analysis~\cite{wang-etal-2018-glue,wang-etal-2019-superglue}.

% data selection
Data selection is a common domain adaptation method. It has been been introduced before neural methods were popular~\cite{moore-lewis-2010-intelligent,axelrod-etal-2011-domain}
and has later been adapted to neural networks~\cite{duh-etal-2013-adaptation,van-der-wees-etal-2017-dynamic,wang-etal-2018-denoising}. Data selection relies on an intermediate classifier which discriminate between in-domain and out-of-domain data. This classifier is trained relying on the small in-domain dataset and the large out-of-domain dataset and is then applied to
the out-of-domain set to identify the examples closest to the targeted domain. Choosing a selection model and the amount of out-of-domain data to select have a strong impact on the effectiveness of the selection methods~\cite{aharoni-goldberg-2020-unsupervised,gururangan-etal-2020-dont}. Our experiments explore these aspects, in addition to the complementarity of selection with fine tuning.

% dynamic/gradual data selection.
Data selection can be performed in multiple rounds, either to gradually restrict the out-of-domain dataset to less and less data~\cite{van-der-wees-etal-2017-dynamic} or to re-evaluate out-of-domain data as pretraining progresses~\cite{wang-etal-2018-denoising}. Data selection can also be performed as a continuous online process~\cite{wang-etal-2018-denoising,wang2021gradientguided,dou-etal-2020-dynamic}.
Our work focus on single round data selection, the most common setting. The benefit of dynamic selection effectiveness has shown to be variable~\cite{wang-etal-2018-denoising} and its use involves defining a complex schedule which is a research topic in itself~\cite{kumar-etal-2019-reinforcement}.

% task selection
Data selection for domain adaptation is also related to data selection for multitask learning. In that case, the out-of-domain dataset is composed of heterogeneous data from different tasks/domains and the training algorithm favor data from some tasks at the 
expense of others~\cite{pmlr-v70-graves17a,wu2020multitask,standley2020multitask}. Contrary to our setting, selection operates only at the task level and the association of training examples to tasks is already known. Multitask learning is an active area of research. This area has explored dynamic selection with reinforcement learning~\cite{pmlr-v70-graves17a,guo-etal-2019-autosem} as well as update projections to align out-of-domain gradients to in-domain gradients~\cite{yu2020surgery,ldery-aux-taks-iclr21}. Some of these ideas have later been investigated in the context of data selection for domain adaptation~\cite{wu-etal-2018-reinforced,kumar-etal-2019-reinforcement,wang2021gradientguided}.

\section{Data Selection Methods}

This section presents the selection method our experiments will focus on and
introduce the trade-offs involved in choosing data selection hyperparameters.

\subsection{In-Domain Data Selection}

Domain adaptation has been introduced for application domains where data reflecting the inference 
conditions is only available in limited quantity. This setting considers that two training sets are available,
a large generic out-of-domain dataset and a small specialized in-domain dataset from the targeted domain~\cite{sogaard2013semi}.
% domain adapt
Classical machine learning assumes that training and test data originate from the same distribution.
At the same time, statistical modeling reaches better generalization performance with large training sets~\cite{vapnik1998statistical}.
Domain adaptation therefore faces a tension between using a large data set with a distribution possibly far from the test conditions and using a small training set matching the test condition.

Data selection tries to address this dilemma. It examines the out-of-domain data and identifies
training examples likely to be most effective at improving the in-domain training loss.
For neural methods, data selection is often used in conjunction with fine tuning in a three phases
process, as shown in Algorithm~\ref{algo}. In a first phase, the model is pretrained on all
the out-of-domain data. In a second phase, an intermediate classifier is trained to distinguish
in-domain from out-of-domain data, using both training sets. The classifier is applied
to the out-of-domain set to identify examples considered close to in-domain data. The intermediate
classifier is then no longer required and the main model is trained on the selected data
starting from the pretrained parameters. Finally, the main model is fine tuned, i.e. it is trained
on the small in-domain training dataset starting from the parameters after the selection phase.

\begin{algorithm}[t]
\SetAlgoLined
 \DontPrintSemicolon

 \KwIn{$D, T$ out and in domain train sets.}
 \KwOut{$\theta$ trained model parameters.}
  ~\\
 \SetKwFunction{FMain}{Select}
  \SetKwProg{Fn}{Function}{:}{}
  \Fn{\FMain{$D$, $T$, $n$}}{
 $w \leftarrow$ trainClassifier($D \cup T$)\;
 $Y \leftarrow$ classify($w, D$)\;
 \KwRet $\argtop_n(Y)$\;
 }
 ~\\~\\
 \SetKwFunction{FMain}{Main}
  \SetKwProg{Fn}{Function}{:}{}
  \Fn{\FMain{$D$, $T$}}{
 $\theta_0 \leftarrow$ initParam()\;
 $\theta_{\rm pre} \leftarrow$ train($\theta_0, D$) ~\texttt{\#pretraining}\;
 $D_{\rm sel} \leftarrow$ select($D, T, n$)\;
 $\theta_{\rm sel} \leftarrow$ train($\theta_{\rm pre}, D_{\rm sel}$)\;
 $\theta_{\rm ft} \leftarrow$ train($\theta_{\rm sel}, T$) ~\texttt{\#fine-tuning}\;
 \KwRet $\theta_{\rm ft}$\;
 }

 \caption{Data Selection \& Fine Tuning for Neural Models\label{algo}}
\end{algorithm}

%The intermediate classification method is an important choice in data selection,
%as is the hyperparameter $n$ setting the number of selected examples.
\noindent\textbf{Contrastive Data Selection:} Commonly, classification is done by estimating 
the probability that a given out-of-domain example $x$ belongs
to the target domain, $P({\cal T}| x)$.
Such an estimation can be done by contrasting the likelihood 
estimated by in-domain LM, $P(\cdot| {\cal T})$ 
and an out-of-domain LM, $P(\cdot | {\cal D})$, i.e.
\begin{equation}
\small
\log P({\cal T}| x) = \log P(x | {\cal T}|) - \log P(x | {\cal D}) + C
\label{eq:cds}
\end{equation}
where $C$ is a constant (log prior ratio). This method was introduced as
{\it intelligent selection}~\cite{moore-lewis-2010-intelligent}
and was later renamed {\it contrastive data selection} (CDS)~\cite{wang-etal-2018-denoising}. 
Initially, it relied on independent n-gram LMs for estimating $P(\cdot| {\cal T})$
and $P(\cdot | {\cal D})$, trained respectively on the (small) in-domain 
training set $T$ and the (large) out-of-domain training set $D$~\cite{moore-lewis-2010-intelligent,axelrod-etal-2011-domain}. 
With neural LMs, $P(\cdot| {\cal T})$ can be estimated by fine-tuning $P(\cdot| {\cal D})$
as suggested by~\cite{van-der-wees-etal-2017-dynamic,wang-etal-2018-denoising}.

The fine tuning strategy is particularly efficient when one performs data selection to
adapt a language model. In that case, there is no need for an intermediate 
model. The pretrained language model to adapt is itself fine-tuned in a few steps
on $T$ and is itself used to score the out-of-domain set. 
\vspace{0.5cm}

\noindent\textbf{Classifier Selection:}
Discriminative classification (DC), introduced by \citet{aharoni-goldberg-2020-unsupervised,jacovi-etal-2021-scalable}, trains a binary classifier to distinguish $T$ and $D$ examples.
This classifier is either trained from scratch or fine tuned from a pretrained model~\cite{devlin-etal-2019-bert,liu2019roberta}. 
\citet{aharoni-goldberg-2020-unsupervised} train the domain classifier, which they refer to as ``Domain-Finetune'', only on the source (English) side of the parallel corpus.
We propose two alternative domain classifiers, that instead condition the classifier on either the target language or both source and target concatenated. 
To finetune language models on the target language data, we use BERT models that are pretrained on German \citep{germanbert}, Russian \citep{kuratov2019adaptation} and multilingual BERT \cite{DBLP:journals/corr/abs-1810-04805}.

The motivation for these alternative classifiers are two fold:
(1) noisy web crawled translation datasets often have incorrect translations (or even languages) which could be missed by the domain classifier if only conditioning on the English source data,
(2) the multilingual domain classifier is able to model the interaction between the source and target and is more analogous to the \textit{bilingual cross-entropy difference} proposed by \citet{axelrod-etal-2011-domain}

Compared to CDS, DC trains a different model which adds training overhead.
On the other hand, a distinct intermediate model offers more flexibility.
The classifier might be pretrained on a different task (e.g. masked LM to select translation data) and its capacity
can be selected independently from the hyperparameter of the model to be adapted. Both aspects are important
since intermediate models can easily overfit given the small size of the target domain set $T$.

\noindent\textbf{Nearest Neighbor Selection:}
A lesser used methods is sentence embedding nearest neighbors~\cite{gururangan-etal-2020-dont,aharoni-goldberg-2020-unsupervised}.
Embedding nearest neighbors relies on a pretrained model~\cite{devlin-etal-2019-bert,liu2019roberta} to represent
sentences as vectors and then measure a domain-score by comparing the distance between a 
candidate sentence vector $v(x)$ and the average in-domain sentence vector $\frac{1}{|T|}\sum_{x \in T} x$. 

In our experiments, we evaluate both constrastive data selection, the most common method by far, and selection with discriminitative classifiers as it has been shown more effective in subsequent work~\cite{aharoni-goldberg-2020-unsupervised}. Previous work and our preliminary experiments indicated that nearest neighbor selection was not competitive with other baselines so we do not include it in our analysis.

% \subsection{Domain Classifier}

% \citet{aharoni-goldberg-2020-unsupervised} propose the use of domain classifiers for data selection.
% They train the domain classifier, which they refer to as ``Domain-Finetune'', only on the source (English) side of the parallel corpus.
% We propose two alternative domain classifiers, that instead condition the classifier on either the target language or both source and target concatenated. 
% To finetune language models on the target language data, we use BERT models that are pretrained on German \citep{germanbert}, Russian \citep{kuratov2019adaptation} and multilingual BERT \cite{DBLP:journals/corr/abs-1810-04805}.

% The motivation for these alternative classifiers are two fold:
% (1) noisy web crawled translation datasets often have incorrect translations (or even languages) which could be missed by the domain classifier if only conditioning on the English source data,
% (2) the multilingual domain classifier is able to model the interaction between the source and target and is more analogous to the \textit{bilingual cross-entropy difference} proposed by \citet{axelrod-etal-2011-domain}
%to adapt \textit{intelligent selection} from \citet{moore-lewis-2010-intelligent} to the translation task. 

\subsection{Hyperparameter Trade-offs}

Data selection for domain adaptation requires selecting several hyperparameters:
the {\it number of pretraining steps}, i.e. when to transition from training on the full out-of-domain set to the selected subset;
the {\it number of selection steps}, i.e. how long to train the model on the selected data; 
the {\it fraction of selected data}, i.e. the size of the selected subset. %If one wants to perform multiple steps of data selection, e.g. gradual data selection~\cite{van-der-wees-etal-2017-dynamic} or curriculum data selection~\cite{kumar-etal-2019-reinforcement}, these parameters much be set for each stage.

These parameters are important as they impact the computational cost of training and
the target domain generalization performance. To examine these trade-offs, the difference between pretraining and fine-tuning is important. Pretraining on a large dataset starts with an initial strong generalization improvement, followed by a long session where the rate of generalization improvement is still positive but ever diminishing. 
%Fine-tuning on a small set after pretraining is different. 
Fine tuning gives a strong generalization improvement in a few steps before overfitting quickly. The fraction of selected data allows trading off between these two extremes: a large fraction of selected data results in a large training set with a distribution close to the out-of-domain distribution while a small fraction results in small training set with a distribution close to the in-domain distribution. This means that settings with large fractions can perform more steps with generalization improvement albeit at a slower pace compared to lower fraction settings. Thus the number of selection steps and the selected fraction parameter interact. Our experiments investigate this interaction.

%When relying on data selection in conjunction with fine tuning, it is important to consider overfitting of the intermediate selection classifier. 
We characterize the effects of overfitting of the intermediate selection classifier, which uniquely affects data selection in conjunction with finetuning.
The intermediate classifier is trained on the small target domain set $T$.
As any machine learning model, it is biased toward its training set and the data it selects can reflect this bias. 
The selected out-of-domain examples might resemble the examples of $T$ more than other in-domain examples unseen during training. 
This bias transferred to the selected data is itself inherited by the model trained on the selected data. 
%To quantify this effect, our experiments measure indirect overfitting and reports the loss on the in-domain training set $T$ during the selection phase. 
This indirect overfitting is crucial for later fine tuning: we report that, in some cases, the selected data is too similar to $T$. There, the complementary value of selection and fine tuning vanishes as data selection fails to identify data providing updates complementary to those provided later by fine tuning on $T$.

\section{Experiments}

We evaluate domain adaptation with data selection on two tasks, language modeling (LM) and machine translation (MT).
For both tasks, we have a large out-of-domain dataset and a small number of examples from the target domain. 
Both sets of data fulfil two functions each.
The out-of-domain data is used to pretrain the model and all the selected data come from the out-of-domain set.
The small target domain set is used to train the intermediate model that scores examples for data selection and, critically, this same set is used for finetuning the final model.
For evaluation, we also have a validation set and test set from the target domain.
The validation set is used to select hyperparameters and early stopping points and the test set is only used for the final model evaluation.

For language modeling, we use the 4.5 million sentences from the One Billion Word corpus \cite{41880} 
as the out-of-domain set and 5k sentences from the Yelp corpus as the target domain. 
This dataset was used for domain adaptation by \citep{oren-etal-2019-distributionally}
and we use their filtered and preprocessed version of the data, including the 1k Yelp validation set and 10k Yelp test set.
We train 2 language models; a 2-layer LSTM recurrent network \cite{zaremba2014recurrent} and a base-size transformer \cite{10.5555/3295222.3295349}.
%We contrast the performance on this smaller model with those on the larger translation model which has significantly more capacity for pretraining.

Our machine translation experiments focus on English-to-German and English-to-Russian.
For the out-of-domain set, we use 4.5 million English-to-German pairs and and 5.2 million English-to-Russian pairs taken from filtered Paracrawl \cite{espla-etal-2019-paracrawl}. 
Paracrawl is composed of translations crawled from the web. Even though we use the filtered version of the dataset, Paracrawl is still noisy including examples of entirely mismatched sentences and occasionally incorrect languages. As in domain data, we rely on news data from the News Commentary Dataset \cite{TIEDEMANN12.463}, which are high quality translations from the news domain. Our in-domain set is limited to 6k sentence pairs.
We use an additional 3k for validation and 10k as the test set.
As a neural MT model, we train
a base transformer \cite{10.5555/3295222.3295349}.
Code to reproduce our experiments is available\footnote{https://git.io/JuAAL}. Models are implemented with Flax \cite{flax2020github}.

We finetune on the small in-domain set by grid searching for a learning rate and using the validation set for early stopping.

\subsection{Selection Methods}

\textbf{Contrastive Data Selection}
The base pretrained (PT) model is fine-tuned (FT) on the small target domain dataset.
This model acts as the ``intermediate'' model in this setting.
Each example in the out-of-domain dataset is scored by the difference between the log likelihoods of the fine-tuned model and the pretrained model.
The full dataset can be ranked by this score and a threshold is selected to train on a uniform distribution of only the top examples.

\noindent \textbf{Discriminative Classifier}
The target domain dataset is used as positive examples and random samples from the out-of-domain dataset are used as negative examples to train a discriminative domain classifier.
The classifier can be a new model trained from random weights, the base model with a binary classification head or a pretrained model from another task (such as a generic masked language model).
Unlike CDS, the base model is not necessarily reused.
The input features to the classifier may either be representations learned from the pretrained base model, other embeddings or the raw text data.
In the case of machine translation, the classifier can be trained on the source, target or both.

%In our LSTM experiments, we evaluate CDS and a discrimative classifier, i.e. a 2-layer feed forward network on the mean of the output embeddings from the LSTM LM.
In our transformer experiments, we evaluate CDS and two classifiers, 
(i) a logistic regression model on bytepair encodings~\cite{sennrich-etal-2016-neural} and (ii) a fine-tuned BERT classifier \cite{germanbert, kuratov2019adaptation, DBLP:journals/corr/abs-1810-04805}. 
We use four settings for the BERT classifier, training on the source, target, mean of the former two, and concatenated language pairs, using the respective language specific pretrained BERT.
For the concatenated case, we use a multilingual BERT.
% This choice utilizes a pretrained model, effectively filters out incorrect language pairs and is sufficient at modeling the domain.

\begin{table}[h!]
\small
\begin{center}
 \begin{tabular}{||l | r r | r r ||}
 \hline
  & \multicolumn{2}{c|}{En-De}  & \multicolumn{2}{c||}{En-Ru}  \\ [0.25ex] 
   & \multicolumn{2}{c|}{{logPPL~~BLEU}}  & \multicolumn{2}{c||}{logPPL~~BLEU}\\
 \hline\hline
PT  & 1.666 & {\it 23.71} & 1.815 & {\it 23.20} \\
+FT & 1.612 & {\it 26.89} & 1.708 & {\it 24.92} \\\hline
PT + CDS & 1.626 & {\it 26.77} & 1.757 & {\it 24.08} \\
+FT  & 1.608 & {\it 27.27} & 1.707 & {\it 25.08} \\ \hline
PT + DC (LogReg)  & 1.624 & {\it 26.22} & 1.762	& {\it 23.43} \\ 
+FT  & 1.575 & {\it 27.54} & 1.666 & \textit{25.35} \\ \hline
PT + DC (BERT)  & 1.599 & {\it 26.33} & 1.752 &	\textit{23.66} \\
+FT  & \textbf{1.550} & \textbf{\textit{27.78}} & \textbf{1.645} & \textbf{\textit{25.52}} \\

\hline
\end{tabular}
\caption{Data selection for machine translation of English to German and English to Russian. BLEU in italics next to log-perplexity (log PPL). For both datasets, models were trained to 200K steps of pretraining and 15k steps of data selection.}
\label{table:nmtresults}
\end{center}
\end{table}

\begin{table}[h!]
\small
\begin{center}
 \begin{tabular}{||l | r r | r r | r ||}
 \hline
  & \multicolumn{2}{c|}{En-De}  & \multicolumn{2}{c|}{En-Ru}  & LM \\ [0.25ex] 
   & \multicolumn{2}{c|}{{lgPPL~~BLEU}}  & \multicolumn{2}{c|}{lgPPL~~BLEU} & lgPPL \\
 \hline\hline
PT  &  1.00 & 	1.00 & 	1.00	 & 1.00 & 1.00\\
+FT & 1.00 & 	1.00 & 	1.00	 & 0.992 & 1.00 \\\hline
CDS & 1.00 & 	1.00 & 	1.00	 & 1.00 & 1.00 \\
+FT  & 1.00 & 	0.998 & 	1.00 & 	0.975  & 1.00 \\\hline
DC-LR  & 1.00 & 	1.00	 & 1.00 & 	1.00 & 1.00 \\
+FT  & 0.951 & 	0.890    & 	0.840 & 	0.742 & 0.998 \\\hline
DC-BERT  & 1.00	 & 1.00	 & 1.00 & 	1.00 & 1.00\\
+FT  & - & - & - & - & -  \\
\hline
\end{tabular}
\caption{Paired bootstrap comparison: each value reports the fraction of samples with worse 
mean performance than PT + DC-BERT + FT for 1k samples of 10k sentences sampled from a 10k 
sample test set.}
\label{table:nmtstatsig}
\end{center}
\end{table}

\subsection{Training on Selected Data}

\noindent\textbf{Machine Translation}
Table~\ref{table:nmtresults} reports the log-perplexity and BLEU scores on two language pairs for each of the selection methods described above.
Data selection always outperforms the baseline without selection, with the BERT domain classifier producing the best log-probability and BLEU on both datasets.
The effectiveness of DC compared to CDS is a surprising result given the popularity of CDS.
We fix the number of training steps on the selected data to 15K and pretrain the baseline model for an additional 15k steps so there is the same number of pretraining + finetuning steps for all settings.
We search the optimal selection size for this cutoff of training steps, which we found to be 1 million for En-Ru and 500k for En-De.
We report results before and after finetuning to highlight the variation in effectiveness of finetuning after the alternative selection methods.
This is particularly noticeable for En-Ru where CDS outperforms the logistic regression classifier before finetuning but is worse after finetuning. 
In all settings, finetuning is more effective after data selection with a discriminative classifier rather than with CDS.
Section~\ref{sec:complement} provides insight as to why this is the case.

Table~\ref{table:nmtstatsig} reports the paired bootstrap resampling \cite{koehn-2004-statistical} where the PT + DC (BERT) + FT model is compared to the baseline models, in terms of loss and BLEU, corresponding to Table~\ref{table:nmtresults}. 
Each value is computed from the 10,000 example test set.
We draw 1,000 bootstrap samples of 10,000 points each, with replacement.
This test shows that the classifier method of data selection outperforms CDS with over 99\% statistical significance on log-perplexity. 

\begin{figure}[h!]
  \includegraphics[trim=75 0 125 50,clip,width=\columnwidth]{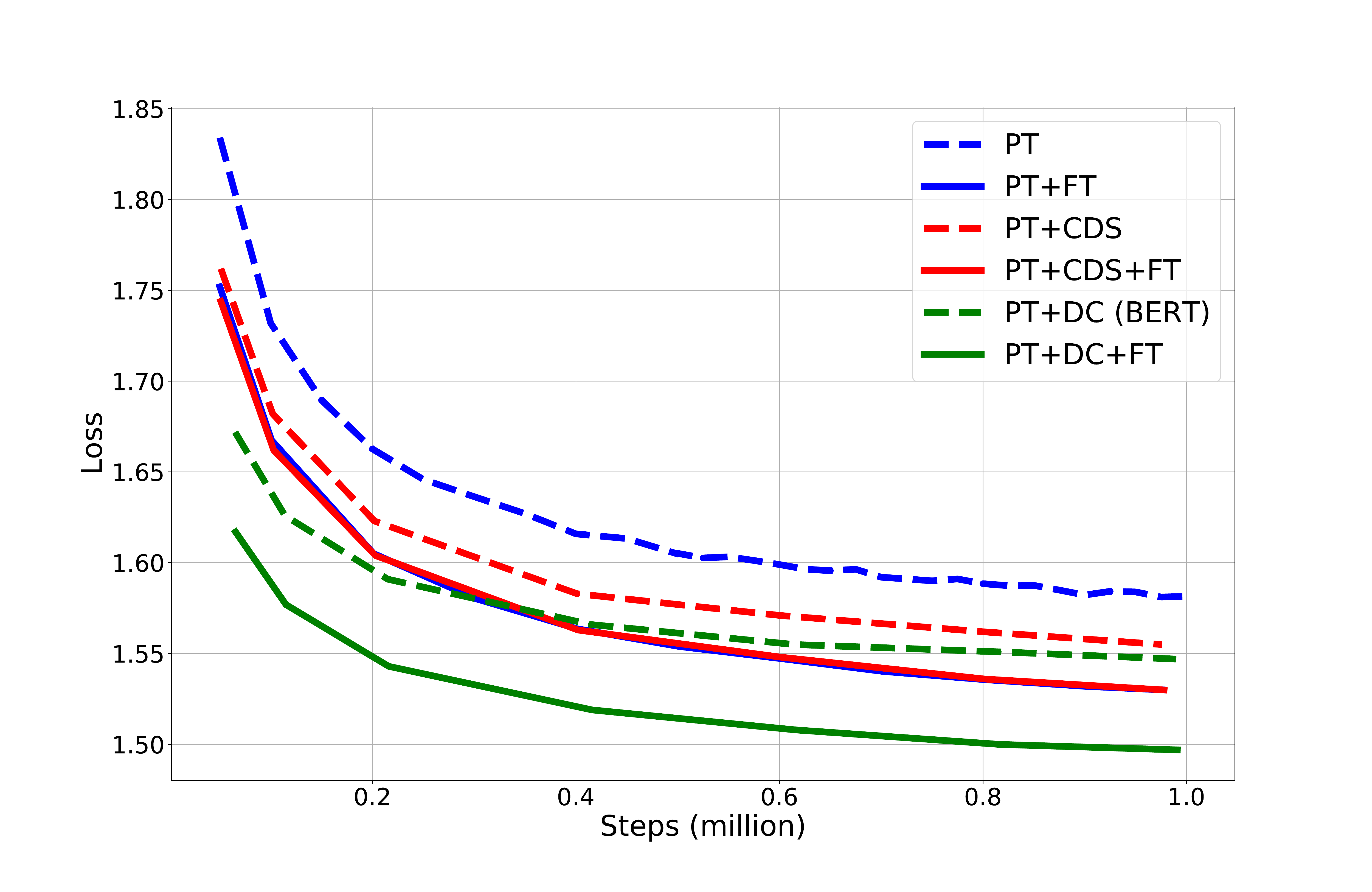}
    \caption{The validation loss curves for pretraining, data selection and finetuning (MT En-De). The pretraining loss (PT) is a single training run, whereas all the other points are checkpoints from the base run that were trained on selected data and/or finetuned.} 
    %The benefits of data selection are consistent throughout pretraining and some of the losses reached with data selection are unreachable with pretraining 1m steps.}
  \label{fig:loss_curve}
\end{figure}

Figure~\ref{fig:loss_curve} shows the log-probabilities at different checkpoints ranging from 50k  to 1 million steps of training.
The relative benefit of FT and DC+FT over PT is diminishing as training progresses. 
However, there are consistent benefits from data selection, so longer pretraining on large models is not sufficient to replace data selection.
Even pretraining up to 1m steps and finetuning (log ppl = 1.530) does not reach the loss from 
DC + FT at 400k (log ppl = 1.519).
The relative improvement between methods is surprisingly constant across pretraining steps with a slight decline 
in the complementary benefit of combining fine tuning with selection. This means that comparing the 
adaptation methods early in the pretraining process is indicative of their relative loss at a later stage.

Further evaluation of performance at different checkpoints throughout pretraining can be found in the Appendix.

% The MT results in Table~\ref{table:nmtresults} and Figure~\ref{fig:loss_curve} show a strong difference between both classifiers (DC) and contrastive selection (CDS), both before and after fine tuning. CDS alone is not beneficial compared to fine tuning but is better than pretraining. CDS in combination with FT barely brings an improvement after 200k steps. 
% On the other end, DC using the BERT classifier alone brings 
% an improvement bigger than FT for some points, while DC with logistic regression outperforms CDS and FT but only after finetuning. The relative benefit of FT and DC+FT over PT is diminishing as training progresses. 
% Even pretraining up to 1m steps and finetuning (log ppl = 1.530) does not reach the loss from 
% DC + FT at 400k (log ppl = 1.519).
% The effectiveness of DC compared to CDS is a surprising result given the popularity of CDS.

\noindent\textbf{Domain Classifier Variants}
Table~\ref{table:clf_results} reports the log-perplexities and BLEU scores for the four variants of the BERT domain classifier.
\citet{aharoni-goldberg-2020-unsupervised} propose the Source DC method.
We propose also exploring target-language-conditioned domain classifiers, and in fact, find that the Target DC selection method outperforms Source DC on En-DE.
Concatenation DC does not yield the best results despite having access to the most data (ie. both source and target).
This may be because of the pretraining mismatch, in that Multilingual BERT was not trained on pairs of segments from different languages.
We also take evaluate using the mean score of the source and target models as a simple alternative to the multilingual BERT approach.
Future work may explore alternative methods for fusing source and target language representations for training a domain classifier.

\begin{table}[h!]
\small
\begin{center}
 \begin{tabular}{||l | r r | r r ||}
 \hline
 & \multicolumn{2}{c|}{En-De}  & \multicolumn{2}{c||}{En-Ru}  \\ [0.25ex] 
& log PPL & BLEU & log PPL & BLEU \\
 \hline\hline
Target DC  & \textbf{1.550} & \textbf{\textit{27.78}} & 1.653 & \textit{25.21} \\
Source DC  & 1.557 & \textit{27.52} & \textbf{1.645} & \textbf{\textit{25.52}} \\
Concat DC  & 1.560 & \textit{27.68} & 1.657 & \textit{25.20} \\
Mean DC & 1.555 & \textit{27.71} & 1.647 & \textit{25.29} \\
\hline
\end{tabular}
\caption{Different types of BERT classifiers, target uses the target language (De/Ru), the source is English and \textit{Concat} concatenates source and target and trains classifier on multilingual BERT. Mean takes the mean scores from source and target classifiers. All models are evaluated at 200k pretraining steps, similar to Table~\ref{table:nmtresults}.}
\label{table:clf_results}
\end{center}
\end{table}

\begin{table}[h!]
\small
\begin{center}
 \begin{tabular}{||l | r  | r  ||}
 \hline
 & LSTM  & Transformer \\ [0.25ex] 
 \hline\hline
PT  & 4.978 & 4.582  \\ 
+FT & 4.284 &  4.145  \\ \hline
PT + CDS & 4.548 &  4.392\\ 
+FT  & 4.183 &  4.151\\ \hline
PT + DC (LogReg)  & 4.644 & 4.456 \\ 
+FT  &  4.183 &  4.108 \\ \hline
PT + DC (LM Hidden)  & 4.603 &  - \\ 
+FT  & \textbf{4.179} & - \\ \hline
PT + DC (BERT)  & - &  4.385 \\ 
+FT  & - & \textbf{4.069} \\

\hline
\end{tabular}
\caption{Language modeling results (log-perplexity) across selection methods for an LSTM and a base-transformer. The LSTM was pretrained for 115k steps and the transformer was trained for 20k steps.}
\label{table:lmresults}
\end{center}
\end{table}

\noindent\textbf{Language Modeling}
For language modeling we evaluate on both a modestly sized LSTM and a base-size transformer.
For the LSTM domain classifier, we reuse the pretrained language model as the feature representation for a simple linear domain classifier (LM Hidden), as a smaller domain classifier seems appropriate given the smaller language model.
We see similar results for the two models despite the large differences in number of parameters, training steps and proximity to convergence.
The LM results in Table~\ref{table:lmresults} show that fine tuning (PT+FT) and data selection (CDS, DC) are improving the pretrained model
on target domain validation data. The benefit of FT alone is generally greater than selection alone 
but both approaches are complementary with the best result obtained with combined approaches (CDS+FT, DC+FT).
When comparing methods we observe that DC is worse than CDS on its own but it is equivalent or  
better in combination with fine tuning (DC+FT vs CDS+FT). This indicates that the methods differ in their 
complementarity with FT and evaluating selection approaches before fine tuning is not sufficient.

\subsection{Overfitting and Complementarity} \label{sec:complement}

\begin{figure*}[!htb]
\minipage{0.32\textwidth}
  \includegraphics[trim=75 50 170 50,clip,width=\linewidth]{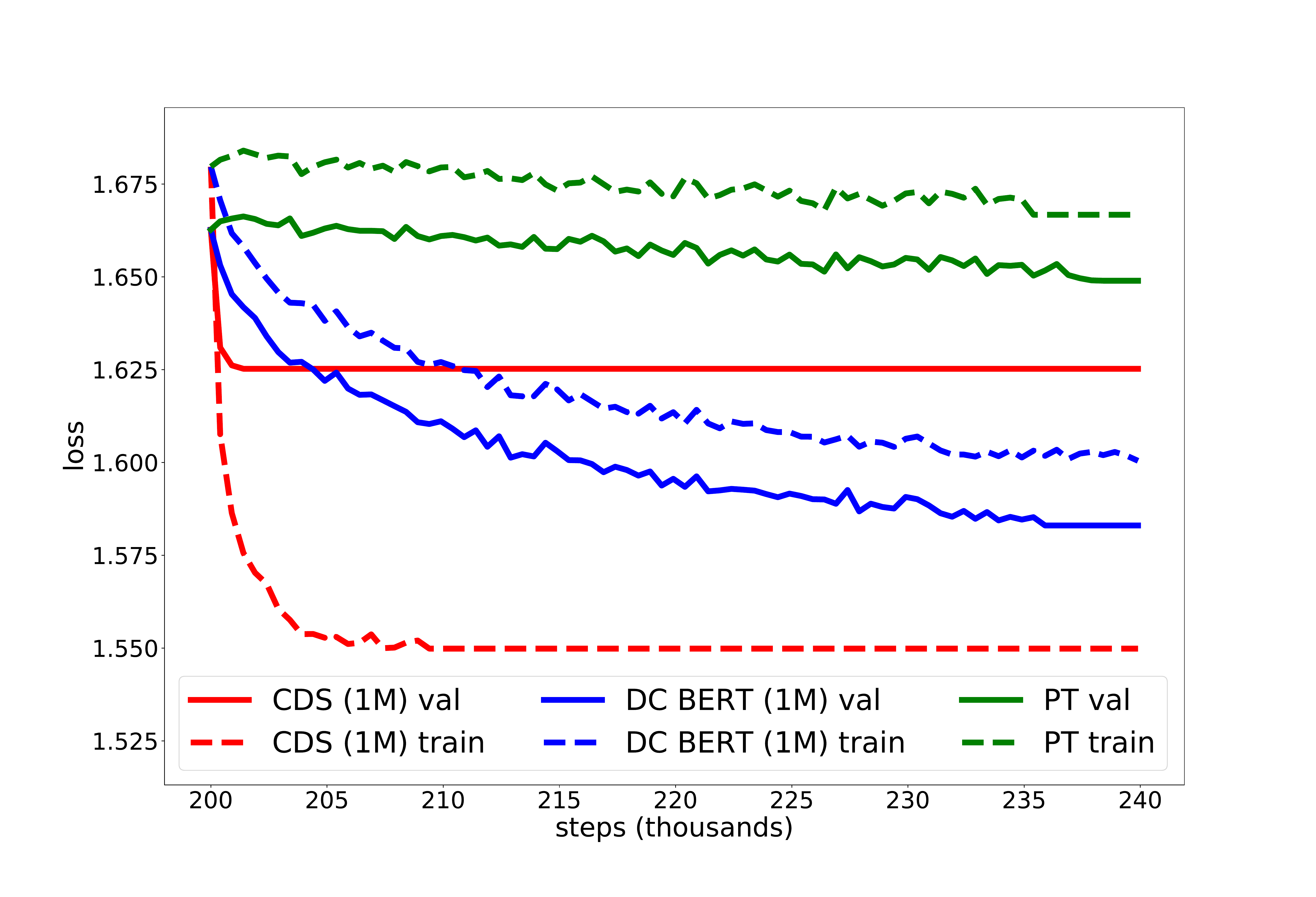}
\endminipage\hfill
\minipage{0.32\textwidth}
  \includegraphics[trim=100 50 125 50,clip,width=\linewidth]{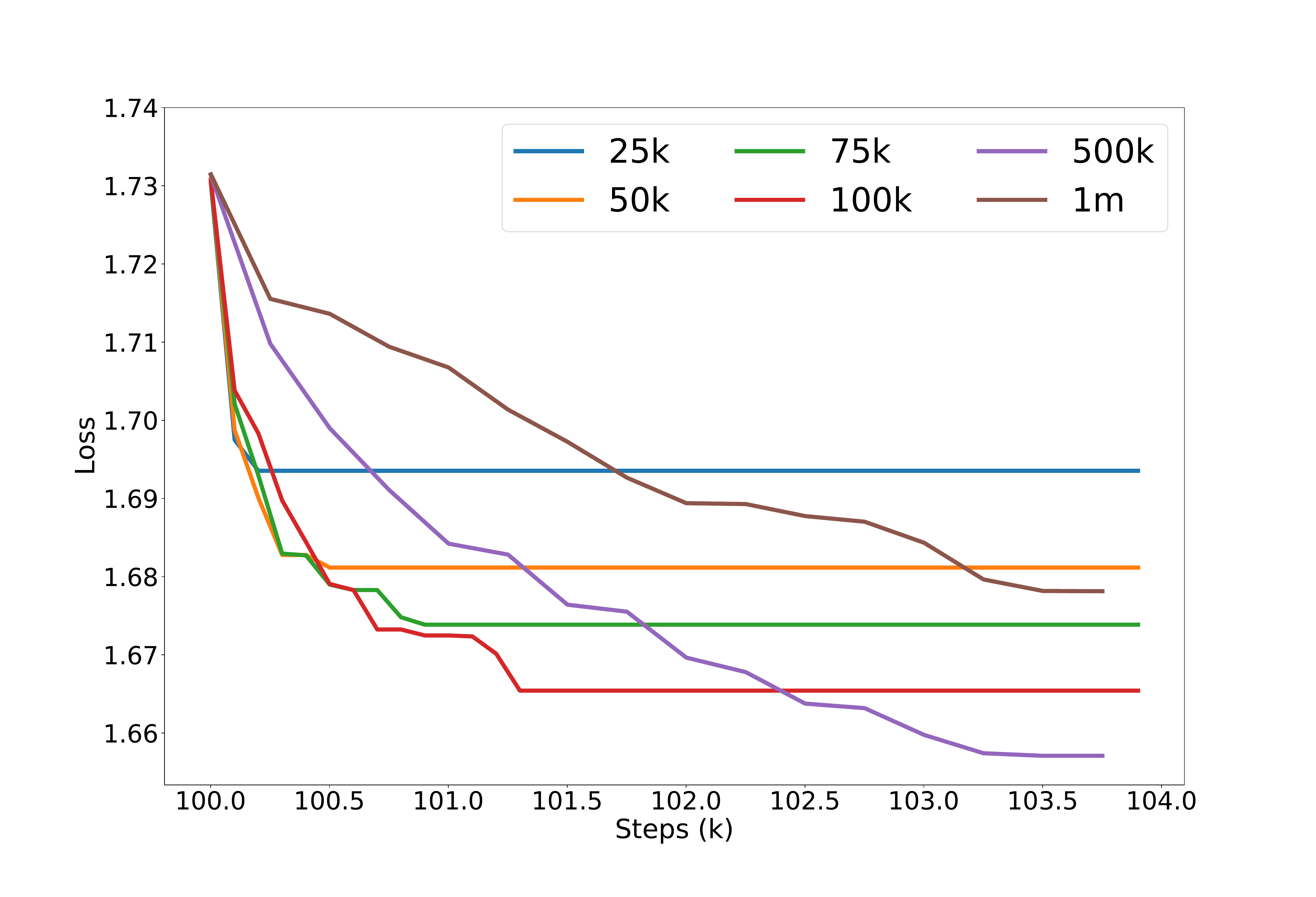}
\endminipage\hfill
\minipage{0.32\textwidth}
  \includegraphics[trim=50 0 100 -50,clip,width=\linewidth]{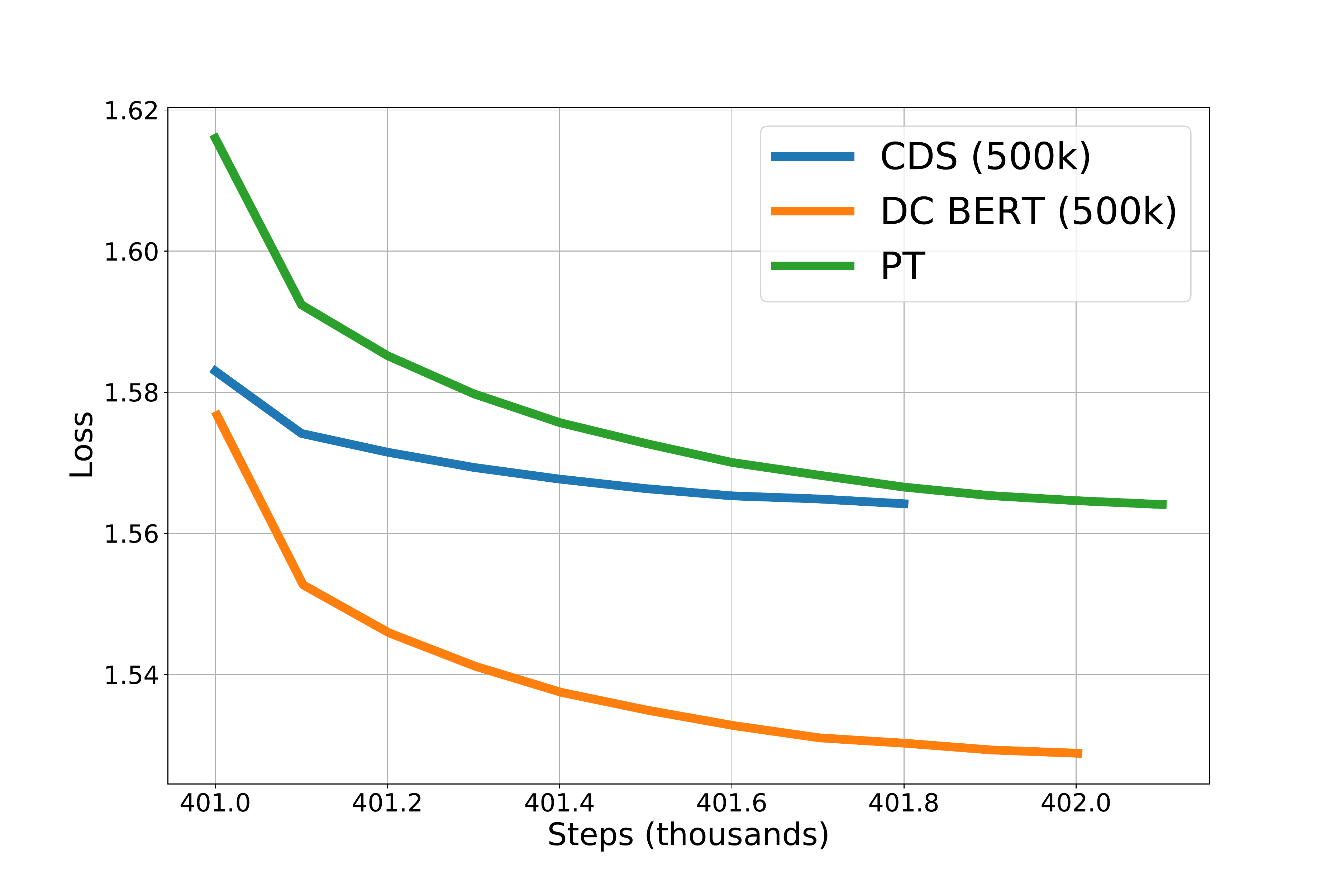}
  \label{fig:finetune}
\endminipage
\vspace{-0.25cm}
\caption{Effects of overfitting and complementarity: \textbf{Left:} Validation and training loss on the target domain during training on selected data  (MT En-De). The dotted line falling below the solid line indicates the model is overfitting to the small target domain dataset despite never seeing this data in training. 
\textbf{Middle:} Loss curves for 6 different data selection sizes for DC (BERT) at the 100k checkpoint (MT En-De). Larger sizes improve loss more slowly but can be trained for longer to eventually outperform the smaller sets. For readability, we display the best checkpoint up to each step.
\textbf{Right:} Validation loss on MT En-De during finetuning. Both data selection methods start at a loss that is better than pretraining but CDS does not benefit much from finetuning, reaching a loss similar to finetuning without data selection. Classifier selection has large a improvement from finetuning.}
\label{fig:overfitting}
\end{figure*}

Our work compares two common data selection techniques, contrastive data selection (CDS) 
and a discriminative domain classifier (DC). As discussed in the previous section, we found
the combination of DC+FT to be the most effective combination both for our LM and MT settings.
One reason of this success is the complementarity of DC with FT. 
CDS did not benefit as much from subsequent fine tuning as DC selection.

% \begin{figure}[h!]
%   \includegraphics[trim=175 50 240 50,clip,width=\columnwidth]{overfitting_long_flat}
%     \caption{Validation and training loss on the target domain during training on selected data  (MT En-De). The selected data is chosen based on the in-domain training data but this data is not seen explicitly by the model. The dotted line falling below the solid line indicates the model is overfitting to the small target domain dataset. The train loss starts slightly higher initially due to variance in the samples.}
%   \label{fig:overfitting}
% \end{figure}
%

In Figure~\ref{fig:overfitting} (left), we show the learning curves for both CDS and DC (BERT) with the same selection size of 1m for MT with 200k steps of pretraining. The red dotted curve show that the CDS model reaches excellent performance on the target-domain training set, but fail to perform as well on the target-domain validation set. This means that the MT model trained on CDS selected data suffers more from overfitting than the MT model trained on DC selected data. 
This is particularly surprising given the large selection size of nearly 1/4th of pretraining data.
The data selected by CDS is too specific to the target-domain training set. This bias also 
certainly explains the worse complementary of CDS and FT, i.e. if CDS selects a training set whose effect is similar to the target-domain training set $T$, the updates from $T$ at fine-tuning are less beneficial.

% \begin{figure}[h!]
%   \includegraphics[trim=100 50 125 50,clip,width=\columnwidth]{thresholds_flat}
%     \caption{Loss curves for 6 different data selection sizes for DC (BERT) at the 100k checkpoint (MT En-De). Larger sizes improve loss more slowly but can be trained for longer to eventually outperform the smaller sets.}
%   \label{fig:thresholds}
% \end{figure}

Lastly, we examine important pitfalls to avoid when comparing selection methods and validating their parameters. Figure~\ref{fig:overfitting} (middle) shows that when considering selection sets of different sizes, training curves converges at different rates. Small selected subsets progress at the fastest rate but reaches their best generalization quickly, and subsequently overfit, while large subsets progress at a slower rate but their best generalization later. This means that short diagnostics to pick the subset size will under estimate the value of large subsets. This is problematic for efficiently defining curriculum with data selection~\cite{kumar-etal-2019-reinforcement}. Similarly, the generalization loss of model which went through a data selection phase but prior to fine tuning is also misleading to predict its loss after fine tuning as illustrated in Figure~\ref{fig:overfitting} (right).

% \begin{figure}[h!]
%   \includegraphics[trim=50 0 100 50,clip,width=\columnwidth]{complementarity_flat}
%     \caption{Validation loss on MT En-De during finetuning after pretraining or data selection (with selection size of 500k). Both data selection methods start at a loss that is better than pretraining but due to overfitting on the training data during data selection, CDS does not improve enough and reaches a best loss similar to pretraining and finetuning for the same number of steps. DC has large a improvement from finetuning.}
%   \label{fig:finetune}
% \end{figure}

\subsection{Effectiveness of Data Selection}

The purpose of the intermediate data selection model is to rank all the out-of-domain data from most to least similar with respect to the in-domain data.
We evaluate and report the performance of CDS and DC for both LM and MT tasks.
The data selection model is never used explicitly as a binary classifier but rather as a scorer.
However, as a proxy for the quality of scoring, we evaluate the binary classification accuracy on an unseen set of in-domain and out-of-domain data.
We also report the average quantile of the in-domain validation data which simulates where in the ranking true in-domain examples would appear.
We split the out-of-domain data into 100 equal bins and take the average of the bin index that each in-domain example would fall into by its data selection score.

Table~\ref{table:selectionacc} shows good performance of CDS and DC for language modeling but clear underperformance of CDS as a binary classifier in the MT setting.
Also, it is noteworthy that logistic regression on byte-pair unigrams outperforms CDS and approaches the performance of BERT while having many fewer parameters and a much lower training cost.

\begin{table}[h!]
\begin{center}
 \begin{tabular}{|| c | l | r r  ||}
 \hline
 &Classifier & Accuracy & Avg Quant. \\ 
 \hline\hline
 \multirow{2}{*}{LM}&  CDS &  91.65\% & 3.6 \\
 & MLP & 89.02\% & 4.9 \\
 \hline\hline
 \multirow{3}{*}{\shortstack[l]{~~~MT~\\(En-De)}} & CDS & 66.94\% & 26.0 \\
 & LogReg & 87.52\% & 3.9 \\
 & BERT & 93.51\% & 2.0 \\
 \hline
\end{tabular}
\caption{Binary classification accuracy of domain classifier and average quantile of in-domain data when binned with ranked out-of-domain data.}
\label{table:selectionacc}
\end{center}
\end{table}

\section{Conclusions}
\label{sec:ccl}

This work explores data selection, a popular method for domain adaption for neural language modeling and neural machine translation. 
Data selection typically divides a training run into three phases: pretraining
on out-of-domain data, training on out-of-domain data selected to resemble target domain data and fine tuning on target domain data. We compare the most common selection methods, contrastive data selection and discriminative 
model classifier and measure their complementarity with fine tuning.

Our experiments motivate several practical recommendations for the practitioner:
(i) pretraining followed by data selection and fine tuning can reach a given generalization loss several time faster in terms of total training steps than pretraining with fine tuning;
(ii) a data selection method should not be evaluated before fine 
tuning since not all methods/parameters bring the same complementary value compared to fine tuning;
(iii) data selection should care about overfitting to the in-domain training set, since this type of overfitting results in selected data very similar to the fine tuning set and impacts the complementarity of data selection and fine tuning;
(iv) longer pretraining runs are always beneficial to later adaptation stages for fine-tuning, 
data selection and their combination but pretraining has diminishing return;
(v) despite the popularity of contrastive data selection, discriminative domain classifiers consistently outperformed this method in our experiments.

\bibliography{anthology,custom}

\begin{thebibliography}{46}
\expandafter\ifx\csname natexlab\endcsname\relax\def\natexlab#1{#1}\fi

\bibitem[{Aharoni and Goldberg(2020)}]{aharoni-goldberg-2020-unsupervised}
Roee Aharoni and Yoav Goldberg. 2020.
\newblock \href {https://doi.org/10.18653/v1/2020.acl-main.692} {Unsupervised
  domain clusters in pretrained language models}.
\newblock In \emph{Proceedings of the 58th Annual Meeting of the Association
  for Computational Linguistics}, pages 7747--7763, Online. Association for
  Computational Linguistics.

\bibitem[{Axelrod et~al.(2011)Axelrod, He, and Gao}]{axelrod-etal-2011-domain}
Amittai Axelrod, Xiaodong He, and Jianfeng Gao. 2011.
\newblock \href {https://www.aclweb.org/anthology/D11-1033} {Domain adaptation
  via pseudo in-domain data selection}.
\newblock In \emph{Proceedings of the 2011 Conference on Empirical Methods in
  Natural Language Processing}, pages 355--362, Edinburgh, Scotland, UK.
  Association for Computational Linguistics.

\bibitem[{Bahdanau et~al.(2015)Bahdanau, Cho, and Bengio}]{bahdanau2014neural}
Dzmitry Bahdanau, Kyunghyun Cho, and Yoshua Bengio. 2015.
\newblock Neural machine translation by jointly learning to align and
  translate.
\newblock In \emph{ICLR}.

\bibitem[{Ba{\~n}{\'o}n et~al.(2020)Ba{\~n}{\'o}n, Chen, Haddow, Heafield,
  Hoang, Espl{\`a}-Gomis, Forcada, Kamran, Kirefu, Koehn, Ortiz~Rojas,
  Pla~Sempere, Ram{\'\i}rez-S{\'a}nchez, Sarr{\'\i}as, Strelec, Thompson,
  Waites, Wiggins, and Zaragoza}]{banon-etal-2020-paracrawl}
Marta Ba{\~n}{\'o}n, Pinzhen Chen, Barry Haddow, Kenneth Heafield, Hieu Hoang,
  Miquel Espl{\`a}-Gomis, Mikel~L. Forcada, Amir Kamran, Faheem Kirefu, Philipp
  Koehn, Sergio Ortiz~Rojas, Leopoldo Pla~Sempere, Gema
  Ram{\'\i}rez-S{\'a}nchez, Elsa Sarr{\'\i}as, Marek Strelec, Brian Thompson,
  William Waites, Dion Wiggins, and Jaume Zaragoza. 2020.
\newblock \href {https://doi.org/10.18653/v1/2020.acl-main.417} {{P}ara{C}rawl:
  Web-scale acquisition of parallel corpora}.
\newblock In \emph{Proceedings of the 58th Annual Meeting of the Association
  for Computational Linguistics}, pages 4555--4567, Online. Association for
  Computational Linguistics.

\bibitem[{Chelba et~al.(2013)Chelba, Mikolov, Schuster, Ge, Brants, Koehn, and
  Robinson}]{41880}
Ciprian Chelba, Tomas Mikolov, Mike Schuster, Qi~Ge, Thorsten Brants, Phillipp
  Koehn, and Tony Robinson. 2013.
\newblock \href {http://arxiv.org/abs/1312.3005} {One billion word benchmark
  for measuring progress in statistical language modeling}.
\newblock Technical report, Google.

\bibitem[{Collobert et~al.(2011)Collobert, Weston, Bottou, Karlen, Kavukcuoglu,
  and Kuksa}]{collobert11}
Ronan Collobert, Jason Weston, L{{\'e}}on Bottou, Michael Karlen, Koray
  Kavukcuoglu, and Pavel Kuksa. 2011.
\newblock \href {http://jmlr.org/papers/v12/collobert11a.html} {Natural
  language processing (almost) from scratch}.
\newblock \emph{Journal of Machine Learning Research}, 12(76):2493--2537.

\bibitem[{Daum{\'e}~{III} and
  Jagarlamudi(2011)}]{daume-iii-jagarlamudi-2011-domain}
Hal Daum{\'e}~{III} and Jagadeesh Jagarlamudi. 2011.
\newblock \href {https://www.aclweb.org/anthology/P11-2071} {Domain adaptation
  for machine translation by mining unseen words}.
\newblock In \emph{Proceedings of the 49th Annual Meeting of the Association
  for Computational Linguistics: Human Language Technologies}, pages 407--412,
  Portland, Oregon, USA. Association for Computational Linguistics.

\bibitem[{deepset.ai()}]{germanbert}
deepset.ai.
\newblock \href {https://deepset.ai/german-bert} {Open sourcing german bert}.
\newblock Https://deepset.ai/german-bert.

\bibitem[{Dery et~al.(2021)Dery, Dauphin, and Grangier}]{ldery-aux-taks-iclr21}
Lucio Dery, Yann Dauphin, and David Grangier. 2021.
\newblock Auxiliary task update decomposition: The good, the bad and the
  neutral.
\newblock In \emph{International Conference on Learning Representation (ICLR)}.

\bibitem[{Devlin et~al.(2018)Devlin, Chang, Lee, and
  Toutanova}]{DBLP:journals/corr/abs-1810-04805}
Jacob Devlin, Ming{-}Wei Chang, Kenton Lee, and Kristina Toutanova. 2018.
\newblock \href {http://arxiv.org/abs/1810.04805} {{BERT:} pre-training of deep
  bidirectional transformers for language understanding}.
\newblock \emph{CoRR}, abs/1810.04805.

\bibitem[{Devlin et~al.(2019)Devlin, Chang, Lee, and
  Toutanova}]{devlin-etal-2019-bert}
Jacob Devlin, Ming-Wei Chang, Kenton Lee, and Kristina Toutanova. 2019.
\newblock \href {https://doi.org/10.18653/v1/N19-1423} {{BERT}: Pre-training of
  deep bidirectional transformers for language understanding}.
\newblock In \emph{Proceedings of the 2019 Conference of the North {A}merican
  Chapter of the Association for Computational Linguistics: Human Language
  Technologies, Volume 1 (Long and Short Papers)}, pages 4171--4186,
  Minneapolis, Minnesota. Association for Computational Linguistics.

\bibitem[{Dou et~al.(2020)Dou, Anastasopoulos, and
  Neubig}]{dou-etal-2020-dynamic}
Zi-Yi Dou, Antonios Anastasopoulos, and Graham Neubig. 2020.
\newblock \href {https://doi.org/10.18653/v1/2020.emnlp-main.475} {Dynamic data
  selection and weighting for iterative back-translation}.
\newblock In \emph{Proceedings of the 2020 Conference on Empirical Methods in
  Natural Language Processing (EMNLP)}, pages 5894--5904, Online. Association
  for Computational Linguistics.

\bibitem[{Duh et~al.(2013)Duh, Neubig, Sudoh, and
  Tsukada}]{duh-etal-2013-adaptation}
Kevin Duh, Graham Neubig, Katsuhito Sudoh, and Hajime Tsukada. 2013.
\newblock \href {https://www.aclweb.org/anthology/P13-2119} {Adaptation data
  selection using neural language models: Experiments in machine translation}.
\newblock In \emph{Proceedings of the 51st Annual Meeting of the Association
  for Computational Linguistics (Volume 2: Short Papers)}, pages 678--683,
  Sofia, Bulgaria. Association for Computational Linguistics.

\bibitem[{Espl{\`a} et~al.(2019)Espl{\`a}, Forcada, Ram{\'\i}rez-S{\'a}nchez,
  and Hoang}]{espla-etal-2019-paracrawl}
Miquel Espl{\`a}, Mikel Forcada, Gema Ram{\'\i}rez-S{\'a}nchez, and Hieu Hoang.
  2019.
\newblock \href {https://www.aclweb.org/anthology/W19-6721} {{P}ara{C}rawl:
  Web-scale parallel corpora for the languages of the {EU}}.
\newblock In \emph{Proceedings of Machine Translation Summit XVII Volume 2:
  Translator, Project and User Tracks}, pages 118--119, Dublin, Ireland.
  European Association for Machine Translation.

\bibitem[{Finkel and Manning(2009)}]{finkel-manning-2009-hierarchical}
Jenny~Rose Finkel and Christopher~D. Manning. 2009.
\newblock \href {https://www.aclweb.org/anthology/N09-1068} {Hierarchical
  {B}ayesian domain adaptation}.
\newblock In \emph{Proceedings of Human Language Technologies: The 2009 Annual
  Conference of the North {A}merican Chapter of the Association for
  Computational Linguistics}, pages 602--610, Boulder, Colorado. Association
  for Computational Linguistics.

\bibitem[{Glorot et~al.(2011)Glorot, Bordes, and Bengio}]{glorot2011domain}
Xavier Glorot, Antoine Bordes, and Yoshua Bengio. 2011.
\newblock Domain adaptation for large-scale sentiment classification: A deep
  learning approach.
\newblock In \emph{ICML}.

\bibitem[{Goldberg(2017)}]{goldberg2017neural}
Yoav Goldberg. 2017.
\newblock Neural network methods for natural language processing.
\newblock \emph{Synthesis lectures on human language technologies},
  10(1):1--309.

\bibitem[{Graves et~al.(2017)Graves, Bellemare, Menick, Munos, and
  Kavukcuoglu}]{pmlr-v70-graves17a}
Alex Graves, Marc~G. Bellemare, Jacob Menick, R{\'e}mi Munos, and Koray
  Kavukcuoglu. 2017.
\newblock \href {http://proceedings.mlr.press/v70/graves17a.html} {Automated
  curriculum learning for neural networks}.
\newblock In \emph{Proceedings of the 34th International Conference on Machine
  Learning}, volume~70 of \emph{Proceedings of Machine Learning Research},
  pages 1311--1320. PMLR.

\bibitem[{Guo et~al.(2019)Guo, Pasunuru, and Bansal}]{guo-etal-2019-autosem}
Han Guo, Ramakanth Pasunuru, and Mohit Bansal. 2019.
\newblock \href {https://doi.org/10.18653/v1/N19-1355} {{A}uto{S}e{M}:
  Automatic task selection and mixing in multi-task learning}.
\newblock In \emph{Proceedings of the 2019 Conference of the North {A}merican
  Chapter of the Association for Computational Linguistics: Human Language
  Technologies, Volume 1 (Long and Short Papers)}, pages 3520--3531,
  Minneapolis, Minnesota. Association for Computational Linguistics.

\bibitem[{Gururangan et~al.(2020)Gururangan, Marasovi{\'c}, Swayamdipta, Lo,
  Beltagy, Downey, and Smith}]{gururangan-etal-2020-dont}
Suchin Gururangan, Ana Marasovi{\'c}, Swabha Swayamdipta, Kyle Lo, Iz~Beltagy,
  Doug Downey, and Noah~A. Smith. 2020.
\newblock \href {https://doi.org/10.18653/v1/2020.acl-main.740} {Don{'}t stop
  pretraining: Adapt language models to domains and tasks}.
\newblock In \emph{Proceedings of the 58th Annual Meeting of the Association
  for Computational Linguistics}, pages 8342--8360, Online. Association for
  Computational Linguistics.

\bibitem[{Heek et~al.(2020)Heek, Levskaya, Oliver, Ritter, Rondepierre,
  Steiner, and van {Z}ee}]{flax2020github}
Jonathan Heek, Anselm Levskaya, Avital Oliver, Marvin Ritter, Bertrand
  Rondepierre, Andreas Steiner, and Marc van {Z}ee. 2020.
\newblock \href {http://github.com/google/flax} {{F}lax: A neural network
  library and ecosystem for {JAX}}.

\bibitem[{Jacovi et~al.(2021)Jacovi, Niu, Goldberg, and
  Sugiyama}]{jacovi-etal-2021-scalable}
Alon Jacovi, Gang Niu, Yoav Goldberg, and Masashi Sugiyama. 2021.
\newblock \href {https://www.aclweb.org/anthology/2021.eacl-main.47} {Scalable
  evaluation and improvement of document set expansion via neural
  positive-unlabeled learning}.
\newblock In \emph{Proceedings of the 16th Conference of the European Chapter
  of the Association for Computational Linguistics: Main Volume}, pages
  581--592, Online. Association for Computational Linguistics.

\bibitem[{Koehn(2004)}]{koehn-2004-statistical}
Philipp Koehn. 2004.
\newblock \href {https://aclanthology.org/W04-3250} {Statistical significance
  tests for machine translation evaluation}.
\newblock In \emph{Proceedings of the 2004 Conference on Empirical Methods in
  Natural Language Processing}, pages 388--395, Barcelona, Spain. Association
  for Computational Linguistics.

\bibitem[{Koehn et~al.(2020)Koehn, Chaudhary, El-Kishky, Goyal, Chen, and
  Guzm{\'a}n}]{koehn-etal-2020-findings}
Philipp Koehn, Vishrav Chaudhary, Ahmed El-Kishky, Naman Goyal, Peng-Jen Chen,
  and Francisco Guzm{\'a}n. 2020.
\newblock \href {https://www.aclweb.org/anthology/2020.wmt-1.78} {Findings of
  the {WMT} 2020 shared task on parallel corpus filtering and alignment}.
\newblock In \emph{Proceedings of the Fifth Conference on Machine Translation},
  pages 726--742, Online. Association for Computational Linguistics.

\bibitem[{Kumar et~al.(2019)Kumar, Foster, Cherry, and
  Krikun}]{kumar-etal-2019-reinforcement}
Gaurav Kumar, George Foster, Colin Cherry, and Maxim Krikun. 2019.
\newblock \href {https://doi.org/10.18653/v1/N19-1208} {Reinforcement learning
  based curriculum optimization for neural machine translation}.
\newblock In \emph{Proceedings of the 2019 Conference of the North {A}merican
  Chapter of the Association for Computational Linguistics: Human Language
  Technologies, Volume 1 (Long and Short Papers)}, pages 2054--2061,
  Minneapolis, Minnesota. Association for Computational Linguistics.

\bibitem[{Kuratov and Arkhipov(2019)}]{kuratov2019adaptation}
Yuri Kuratov and Mikhail Arkhipov. 2019.
\newblock Adaptation of deep bidirectional multilingual transformers for
  russian language.
\newblock \emph{arXiv preprint arXiv:1905.07213}.

\bibitem[{Liu et~al.(2019)Liu, Ott, Goyal, Du, Joshi, Chen, Levy, Lewis,
  Zettlemoyer, and Stoyanov}]{liu2019roberta}
Yinhan Liu, Myle Ott, Naman Goyal, Jingfei Du, Mandar Joshi, Danqi Chen, Omer
  Levy, Mike Lewis, Luke Zettlemoyer, and Veselin Stoyanov. 2019.
\newblock \href {http://arxiv.org/abs/1907.11692} {Roberta: A robustly
  optimized bert pretraining approach}.

\bibitem[{Moore and Lewis(2010)}]{moore-lewis-2010-intelligent}
Robert~C. Moore and William Lewis. 2010.
\newblock \href {https://www.aclweb.org/anthology/P10-2041} {Intelligent
  selection of language model training data}.
\newblock In \emph{Proceedings of the {ACL} 2010 Conference Short Papers},
  pages 220--224, Uppsala, Sweden. Association for Computational Linguistics.

\bibitem[{Oren et~al.(2019)Oren, Sagawa, Hashimoto, and
  Liang}]{oren-etal-2019-distributionally}
Yonatan Oren, Shiori Sagawa, Tatsunori Hashimoto, and Percy Liang. 2019.
\newblock \href {https://doi.org/10.18653/v1/D19-1432} {Distributionally robust
  language modeling}.
\newblock In \emph{Proceedings of the 2019 Conference on Empirical Methods in
  Natural Language Processing and the 9th International Joint Conference on
  Natural Language Processing (EMNLP-IJCNLP)}, pages 4227--4237, Hong Kong,
  China. Association for Computational Linguistics.

\bibitem[{Peters et~al.(2018)Peters, Neumann, Iyyer, Gardner, Clark, Lee, and
  Zettlemoyer}]{peters-etal-2018-deep}
Matthew Peters, Mark Neumann, Mohit Iyyer, Matt Gardner, Christopher Clark,
  Kenton Lee, and Luke Zettlemoyer. 2018.
\newblock \href {https://doi.org/10.18653/v1/N18-1202} {Deep contextualized
  word representations}.
\newblock In \emph{Proceedings of the 2018 Conference of the North {A}merican
  Chapter of the Association for Computational Linguistics: Human Language
  Technologies, Volume 1 (Long Papers)}, pages 2227--2237, New Orleans,
  Louisiana. Association for Computational Linguistics.

\bibitem[{Raffel et~al.(2020)Raffel, Shazeer, Roberts, Lee, Narang, Matena,
  Zhou, Li, and Liu}]{raffel-t5-2020}
Colin Raffel, Noam Shazeer, Adam Roberts, Katherine Lee, Sharan Narang, Michael
  Matena, Yanqi Zhou, Wei Li, and Peter~J. Liu. 2020.
\newblock \href {http://jmlr.org/papers/v21/20-074.html} {Exploring the limits
  of transfer learning with a unified text-to-text transformer}.
\newblock \emph{Journal of Machine Learning Research}, 21(140):1--67.

\bibitem[{Sennrich et~al.(2016)Sennrich, Haddow, and
  Birch}]{sennrich-etal-2016-neural}
Rico Sennrich, Barry Haddow, and Alexandra Birch. 2016.
\newblock \href {https://doi.org/10.18653/v1/P16-1162} {Neural machine
  translation of rare words with subword units}.
\newblock In \emph{Proceedings of the 54th Annual Meeting of the Association
  for Computational Linguistics (Volume 1: Long Papers)}, pages 1715--1725,
  Berlin, Germany. Association for Computational Linguistics.

\bibitem[{S{\o}gaard(2013)}]{sogaard2013semi}
Anders S{\o}gaard. 2013.
\newblock Semi-supervised learning and domain adaptation in natural language
  processing.
\newblock \emph{Synthesis Lectures on Human Language Technologies},
  6(2):1--103.

\bibitem[{Standley et~al.(2020)Standley, Zamir, Chen, Guibas, Malik, and
  Savarese}]{standley2020multitask}
Trevor Standley, Amir~Roshan Zamir, Dawn Chen, Leonidas~J. Guibas, Jitendra
  Malik, and Silvio Savarese. 2020.
\newblock \href {http://proceedings.mlr.press/v119/standley20a.html} {Which
  tasks should be learned together in multi-task learning?}
\newblock In \emph{Proceedings of the 37th International Conference on Machine
  Learning, {ICML} 2020, 13-18 July 2020, Virtual Event}, volume 119 of
  \emph{Proceedings of Machine Learning Research}, pages 9120--9132. {PMLR}.

\bibitem[{Tan et~al.(2009)Tan, Cheng, Wang, and Xu}]{tan2009adapting}
Songbo Tan, Xueqi Cheng, Yuefen Wang, and Hongbo Xu. 2009.
\newblock Adapting naive bayes to domain adaptation for sentiment analysis.
\newblock In \emph{European Conference on Information Retrieval}, pages
  337--349. Springer.

\bibitem[{Tiedemann(2012)}]{TIEDEMANN12.463}
Jörg Tiedemann. 2012.
\newblock Parallel data, tools and interfaces in opus.
\newblock In \emph{Proceedings of the Eight International Conference on
  Language Resources and Evaluation (LREC'12)}, Istanbul, Turkey. European
  Language Resources Association (ELRA).

\bibitem[{van~der Wees et~al.(2017{\natexlab{a}})van~der Wees, Bisazza, and
  Monz}]{vanderwees2017dynamic}
Marlies van~der Wees, Arianna Bisazza, and Christof Monz. 2017{\natexlab{a}}.
\newblock Dynamic data selection for neural machine translation.
\newblock In \emph{Proceedings of the 2017 Conference on Empirical Methods in
  Natural Language Processing}, pages 1400--1410.

\bibitem[{van~der Wees et~al.(2017{\natexlab{b}})van~der Wees, Bisazza, and
  Monz}]{van-der-wees-etal-2017-dynamic}
Marlies van~der Wees, Arianna Bisazza, and Christof Monz. 2017{\natexlab{b}}.
\newblock \href {https://doi.org/10.18653/v1/D17-1147} {Dynamic data selection
  for neural machine translation}.
\newblock In \emph{Proceedings of the 2017 Conference on Empirical Methods in
  Natural Language Processing}, pages 1400--1410, Copenhagen, Denmark.
  Association for Computational Linguistics.

\bibitem[{Vapnik(1998)}]{vapnik1998statistical}
V.N. Vapnik. 1998.
\newblock \emph{Statistical Learning Theory}.
\newblock A Wiley-Interscience publication. Wiley.

\bibitem[{Vaswani et~al.(2017)Vaswani, Shazeer, Parmar, Uszkoreit, Jones,
  Gomez, Kaiser, and Polosukhin}]{10.5555/3295222.3295349}
Ashish Vaswani, Noam Shazeer, Niki Parmar, Jakob Uszkoreit, Llion Jones,
  Aidan~N. Gomez, undefinedukasz Kaiser, and Illia Polosukhin. 2017.
\newblock Attention is all you need.
\newblock In \emph{Proceedings of the 31st International Conference on Neural
  Information Processing Systems}, NIPS'17, page 6000–6010, Red Hook, NY,
  USA. Curran Associates Inc.

\bibitem[{Wang et~al.(2018)Wang, Watanabe, Hughes, Nakagawa, and
  Chelba}]{wang-etal-2018-denoising}
Wei Wang, Taro Watanabe, Macduff Hughes, Tetsuji Nakagawa, and Ciprian Chelba.
  2018.
\newblock \href {https://doi.org/10.18653/v1/W18-6314} {Denoising neural
  machine translation training with trusted data and online data selection}.
\newblock In \emph{Proceedings of the Third Conference on Machine Translation:
  Research Papers}, pages 133--143, Brussels, Belgium. Association for
  Computational Linguistics.

\bibitem[{Wang et~al.(2021)Wang, Bapna, Johnson, and
  Firat}]{wang2021gradientguided}
Xinyi Wang, Ankur Bapna, Melvin Johnson, and Orhan Firat. 2021.
\newblock \href {http://arxiv.org/abs/2102.13549} {Gradient-guided loss masking
  for neural machine translation}.

\bibitem[{Wu et~al.(2018)Wu, Li, and Wang}]{wu-etal-2018-reinforced}
Jiawei Wu, Lei Li, and William~Yang Wang. 2018.
\newblock \href {https://doi.org/10.18653/v1/N18-1113} {Reinforced
  co-training}.
\newblock In \emph{Proceedings of the 2018 Conference of the North {A}merican
  Chapter of the Association for Computational Linguistics: Human Language
  Technologies, Volume 1 (Long Papers)}, pages 1252--1262, New Orleans,
  Louisiana. Association for Computational Linguistics.

\bibitem[{Wu et~al.(2020)Wu, Zhang, and R{\'{e}}}]{wu2020multitask}
Sen Wu, Hongyang~R. Zhang, and Christopher R{\'{e}}. 2020.
\newblock \href {https://openreview.net/forum?id=SylzhkBtDB} {Understanding and
  improving information transfer in multi-task learning}.
\newblock In \emph{8th International Conference on Learning Representations,
  {ICLR} 2020, Addis Ababa, Ethiopia, April 26-30, 2020}. OpenReview.net.

\bibitem[{Yu et~al.(2020)Yu, Kumar, Gupta, Levine, Hausman, and
  Finn}]{yu2020surgery}
Tianhe Yu, Saurabh Kumar, Abhishek Gupta, Sergey Levine, Karol Hausman, and
  Chelsea Finn. 2020.
\newblock \href
  {https://proceedings.neurips.cc/paper/2020/file/3fe78a8acf5fda99de95303940a2420c-Paper.pdf}
  {Gradient surgery for multi-task learning}.
\newblock In \emph{Advances in Neural Information Processing Systems},
  volume~33, pages 5824--5836. Curran Associates, Inc.

\bibitem[{Zaremba et~al.(2014)Zaremba, Sutskever, and
  Vinyals}]{zaremba2014recurrent}
Wojciech Zaremba, Ilya Sutskever, and Oriol Vinyals. 2014.
\newblock Recurrent neural network regularization.
\newblock \emph{arXiv preprint arXiv:1409.2329}.

\end{thebibliography}
\bibliographystyle{acl_natbib}

\appendix

\section{Appendix}
\label{sec:appendix}

\begin{figure}[h!]
  \includegraphics[trim=50 0 125 50,clip,width=\columnwidth]{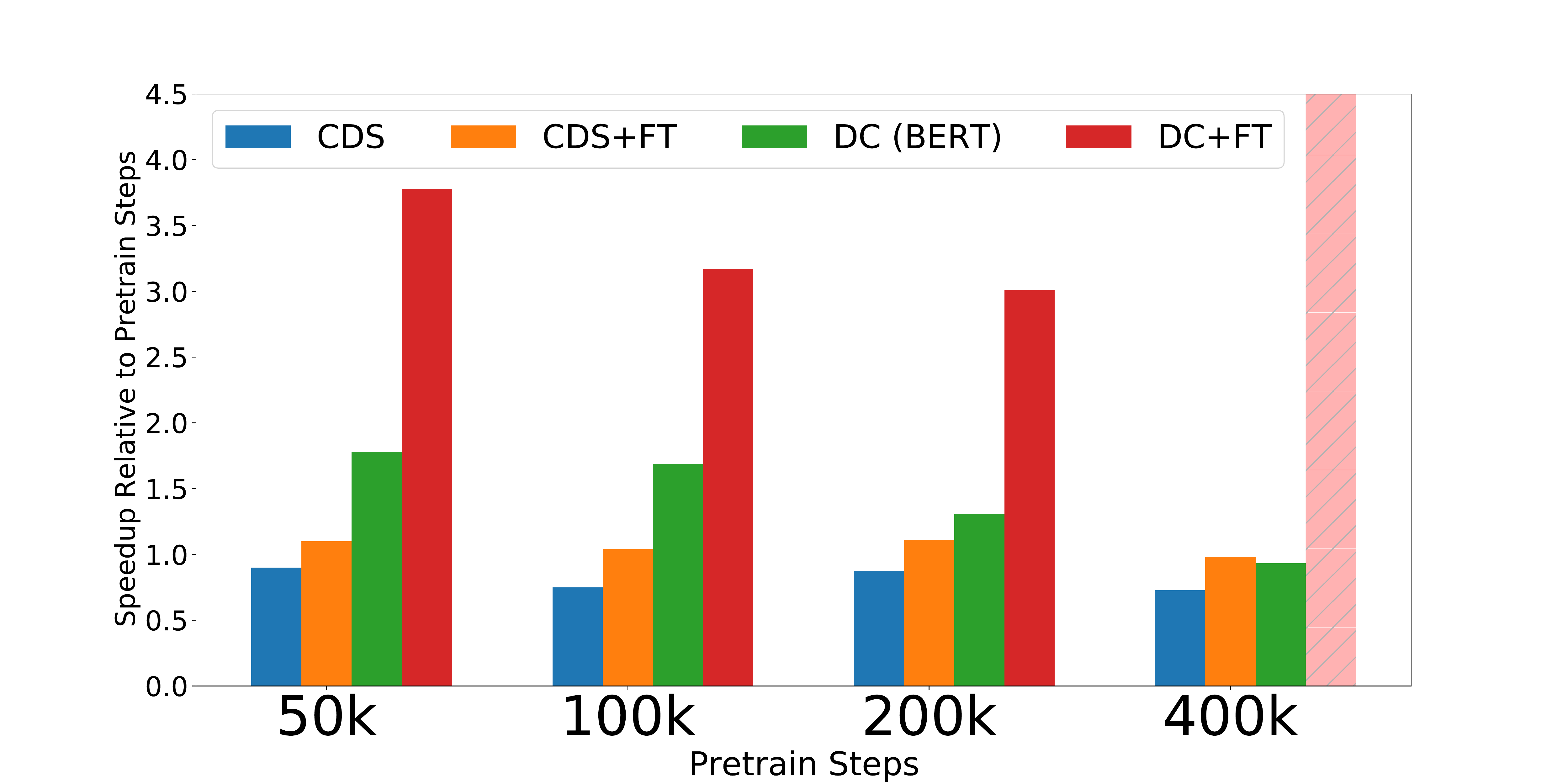}
     \caption{Data selection (MT En-De) as an acceleration method. This table shows the speedup of reaching a given loss at each checkpoint relative to how many steps of pretraining and finetuning are required to reach the same loss. Values lower than 1 indicate that the loss can be reached in fewer steps without data selection. The final bar for DC is shaded to indicate extrapolation and is off the y-axis because the loss is lower than any loss reachable in 1 million steps with pretraining and finetuning.}
   \label{fig:acceration_bars}
\end{figure}

\subsection{Training Steps}
Figure~\ref{fig:acceration_bars} shows the acceleration of training as a 
function of pretraining + finetuning (PT+FT) steps needed to reach an equivalent loss for translation.
This figure highlights the effectiveness of pretraining since the performance obtained by data selection 
for early checkpoints can be matched by simply pretraining longer. 
% In particular, data selection also requires additional exploration of optimal threshold selection, possibly making longer pretraining favorable over selection unless there is a significant improvement from data selection, which is only the case with DC selection for MT in our experiments.
Furthermore, DC+FT at 400k 
pretraining steps cannot be matched, even when pretraining for up to 1m steps. This figure shows that a practitioner with a given generalization requirement can consider data selection early since the target domain generalization gain for early checkpoints might avoid a long pretraining run.

At 50k steps, data selection accelerates training by a factor of about 3.5x, meaning the same performance can be reached with an additional 150k steps of pretraining. However, for later checkpoints, the marginal benefits of pretraining decreases while the improvements from data selection are steady making data selection a clear choice for later checkpoints. In particular for well trained smaller models, such as the LSTM we evaluate for language modeling, the performance after data selection may actually be unreachable just through pretraining either due to the noisiness of the training data that might be filtered from data selection or due to the limited model capacity.

\subsection{Complementary Finetuning vs Overfitting}

\begin{figure}[h!]
  \includegraphics[trim=100 50 100 50,clip,width=\columnwidth]{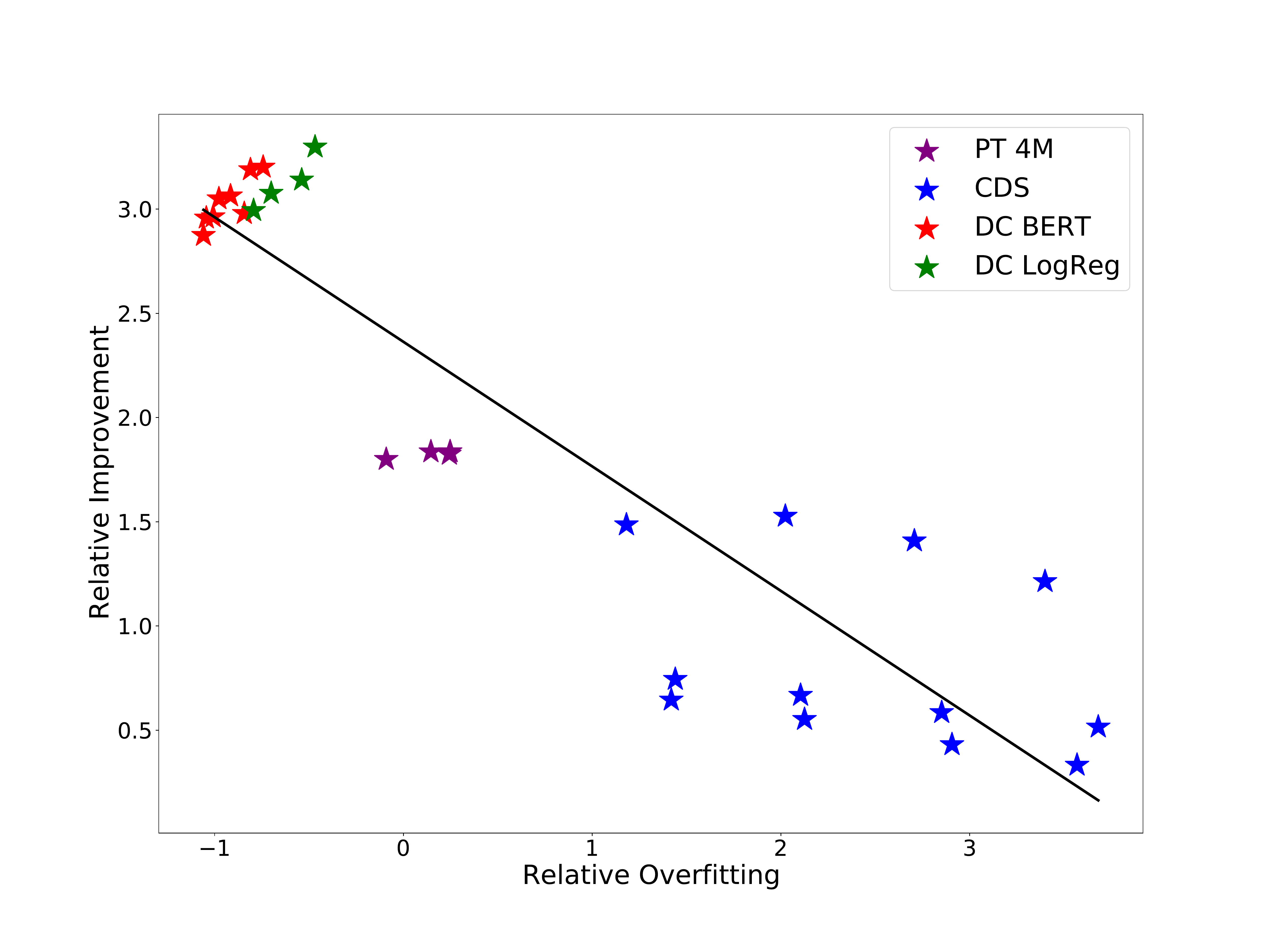}
    \caption{Impact of selection overfitting (MT En-De). When data selection overfits to the in domain set, the improvements from finetuning are lower. The x-axis is the overfitting relative difference and the y-axis is the relative improvement from finetuning. Pearson Correlation Coefficient : -0.91}
  \label{fig:correlate}
\end{figure}

Figure \ref{fig:correlate} measures the correlation between the relative difference between the train and valid best in-domain loss
prior to fine tuning (selection overfitting rate) and the relative difference between the valid loss before and after fine tuning (fine tuning rate). 
There is a strong anti-correlation between these factors, showing that overfitting at the selection stage indeed impacts negatively the impact of FT.
We include points on this graph selecting the top 4m examples, effectively filtering out the bottom 500k, which has a slight overfitting effect, to include more points with an intermediate overfitting-to-improvement tradeoff.

\end{document}